\documentclass{article}

\usepackage{PRIMEarxiv}
\usepackage{amsfonts}       
\usepackage{amsmath}
\usepackage{amssymb}		
\usepackage{bm}
\usepackage{cleveref}
\usepackage{enumitem}		
\usepackage{nicefrac}       
\usepackage{microtype}      
\usepackage{lipsum}
\usepackage{fancyhdr}       
\usepackage{graphicx}		
\usepackage{subcaption}		
\usepackage{booktabs}		
\usepackage{tabularx} 		
\usepackage{multirow}
\graphicspath{{images/}}

\crefname{equation}{Eq.}{Eqs.} 
\Crefname{equation}{Equation}{Equations}

\title{Enhanced forecasting of stock prices based on variational mode decomposition, PatchTST, and adaptive scale-weighted layer}
	
\author{
  Xiaorui Xue \\
  School of Cyber Science and Engineering \\
  Southeast University \\
  Nanjing\\
  \texttt{xuexiaorui@seu.edu.cn} \\
    \And
  Shaofang Li\\
  School of Economics and Management \\
  Southeast Universit \\
  Nanjing \\
  \texttt{shaofangli2003@gmail.com} \\
    \And
  Xiaonan Wang\\
  School of Economics and Management \\
  Southeast Universit \\
  Nanjing \\
  \texttt{beigu806@gmail.com} \\
}

\begin{document}
\maketitle	
    
    \begin{abstract}
        The significant fluctuations in stock index prices in recent years highlight the critical need for accurate forecasting to guide investment and financial strategies. This study introduces a novel composite forecasting framework that integrates variational mode decomposition (VMD), PatchTST, and adaptive scale-weighted layer (ASWL) to address these challenges. Utilizing datasets of four major stock indices—SP500, DJI, SSEC, and FTSE—from 2000 to 2024, the proposed method first decomposes the raw price series into intrinsic mode functions (IMFs) using VMD. Each IMF is then modeled with PatchTST to capture temporal patterns effectively. The ASWL module is applied to incorporate scale information, enhancing prediction accuracy. The final forecast is derived by aggregating predictions from all IMFs. The VMD-PatchTST-ASWL framework demonstrates significant improvements in forecasting accuracy compared to traditional models, showing robust performance across different indices. This innovative approach provides a powerful tool for stock index price forecasting, with potential applications in various financial analysis and investment decision-making contexts. 
    \end{abstract}


\section{Introduction}
Stock price indices serve as barometers for the overall health of financial markets and the economy~\cite{poon2003forecasting}. Accurate forecasting of stock index prices is essential for guiding investment decisions and financial strategies. Given the high-frequency trading and vast number of transactions that occur in stock markets, these systems often exhibit nonlinear and complex behaviors~\cite{math11132883}. Precise predictions of stock index prices can help mitigate the risks of significant financial losses for investors and enable more informed decision-making by financial analysts and policymakers. Thus, there is a critical need for developing a stable, accurate, and broadly applicable forecasting model for stock index prices to effectively navigate market uncertainties.

Despite the complexity of stock index prices, many scholars have explored their predictability from various perspectives. Rapach and Mark examine the predictability of stock prices through valuation ratios, confirming that this factor significantly enhances long-term stock price predictability \cite{rapach2005valuation}. Kim's results validate the feasibility and effectiveness of data-driven models, such as support vector machines (SVM), for short-term stock price predictability \cite{kim2003financial}. Beyond the forecasting horizon, Stock price ratios and return dispersion are used to predict stock returns, demonstrating the enhancement of stock price predictability by these factors \cite{mcmillan2019predicting}. The work by Tiwari et al. indicate that wavelet decomposition strengthens stock price predictability~\cite{tiwari2018output}. Liu et al.'s research further confirm this view \cite{liu2024stock}. All these studies underscore the predictability of stock prices.

There are two main approaches to stock price prediction: multivariate and univariate forecasting. Multivariate forecasting utilizes different types of indicators, such as macroeconomic indicators, policy indicators, technical indicators, and historical stock prices to make predictions~\cite{wang2020forecasting, rapach2005macro, neely2014forecasting, bakas2019volatility}. Recently, textual information has also been incorporated as input variables in several studies to improve the accuracy of stock price predictions~\cite{salisu2020predicting}. One study shows that multivariate stock price forecasting encompasses numerous factors that might influence stock prices, making it more suitable for long-term and overall trend predictions. However, for short-term predictions, an excessive number of variables can introduce more noise, potentially degrading the model's performance~\cite{sezer2020financial}. On the other hand, univariate forecasting uses historical stock price data as the sole input, ignoring the complex indicators involved in multivariate forecasting~\cite{huang2021new, wang2018oil}. This approach implicitly assumes that the effects of all factors are reflected in the price changes. Therefore, it is more suitable for short-term stock price predictions. Univariate forecasting is also more suitable for data-driven models, as these models can extract high-dimensional features from the data, compared to traditional models~\cite{thakkar2021comprehensive}. Based on this, this paper focuses on predicting the price of a single stock index.

Univariate stock price forecasting can be broadly categorized into two types: traditional statistical approaches and data-driven approaches. Initially, traditional methods such as exponential smoothing, autoregressive integrated moving average (ARIMA), and autoregressive conditional heteroskedasticity (ARCH) were widely used for forecasting stock prices and volatility. However, these methods often rely heavily on strict assumptions, rendering predictions for non-stationary stock index price series unreliable~\cite{januschowski2019classification}. To address this limitation, data-driven approaches, including machine learning and deep learning, have been considered. Among machine learning methods, support vector machines (SVMs)~\cite{pai2005hybrid}, decision trees~\cite{basak2019predicting} have demonstrated superior performance compared to traditional approaches in handling non-linearity, high-dimensional data, and small sample sizes. Furthermore, deep learning models inspired by biological systems have gained prominence in financial forecasting. Convolutional neural networks (CNNs), designed to mimic the structure of the neural system~\cite{chen2021novel}, recurrent neural networks (RNNs), which capture temporal dependencies, are among the most notable of these models~\cite{bukhari2020fractional}, and attention-based neural networks, which focus on important parts of the input sequence, are key approaches in modern neural network design. Building upon these foundations, Transformer-based models have further revolutionized time series forecasting. These models leverage self-attention mechanisms to capture long-range dependencies and complex temporal patterns~\cite{vaswani2017attention}. Notable advancements include Informer, which enhances efficiency for long sequences~\cite{zhou2021informer}, Autoformer, which addresses seasonality and trends through decomposition~\cite{wu2021autoformer}, Non-stationary Transformers, which explore the impact of non-stationarity on forecasting accuracy~\cite{liu2022non}, and PatchTST, which segments time series data into patches to capture intricate temporal patterns and improve long-term forecasting performance~\cite{nie2023a}. These innovations underscore the Transformers' growing importance in improving predictive performance for time series data. Notably, the patching approach in PatchTST is particularly well-suited for stock price prediction, as it effectively captures the sequential features across different time steps. Consequently, this study employs PatchTST as the forecasting model.

Due to the dynamic nature of stock prices, it is reported that, despite the impressive performance of Transformer-based models in time series forecasting, achieving satisfactory prediction accuracy directly from individual models remains a significant challenge~\cite{guo2012multi}. To address this issue, a composite forecasting framework combining machine learning/deep learning with decomposition-integration techniques has been proposed. The decomposition-integration techniques decompose the original time series data into multiple independent subsequences, which are then individually predicted and aggregated to obtain the final prediction. Independent component analysis (ICA)~\cite{lee1998independent}, wavelet decomposition, and empirical mode decomposition (EMD)~\cite{huang1998empirical}, along with their variants~\cite{torres2011complete}, have been used to decompose time sequences. These decomposition methods aim to break down complex time series into simpler subsequences of different frequencies and have demonstrated their effectiveness in some price forecasting scenarios~\cite{zhang2015novel, wang2016wind, jianwei2019novel}. Recently, a novel method called variational mode decomposition (VMD)~\cite{dragomiretskiy2013variational} has shown exceptional capabilities in decomposition and feature representation. Particularly in fields such as wind speed forecasting, energy price forecasting, and load forecasting, VMD has exhibited superior predictive performance compared to other decomposition methods~\cite{zhang2023oil, li2021forecasting, wu2022interpretable}. Liu et al. combined VMD with long short-term memory (LSTM) networks to predict non-ferrous metal prices and highlighted the effectiveness of VMD by comparing it with six other state-of-the-art methods~\cite{liu2020non}. Moreover, an integrated stock prediction model consisting of VMD, Extreme Learning Machine (ELM), and Improved Harmony Search (IHS) algorithm significantly improved the accuracy and stability of stock price forecasting~\cite{jiang2022two}.

In general, most composite methods predict multiple sub-sequences directly, meaning that the loss function of deep models used during the training process is the average of the losses from these sub-sequences. However, these sub-sequences often have different scale ranges, which are standardized to a range of 0 to 1 before being input into the model. Consequently, the model treats the sub-sequences equally because it cannot utilize this information, potentially leading to a focus on high-frequency sub-sequences while relatively neglecting more important low-frequency sequences. To address this issue, this paper proposes a novel composite forecasting framework that combines VMD, PatchTST, and adaptive scaled-weighted layer (ASWL), namely VMD+PatchTST with ASWL. The detailed steps of this proposed framework are as follows: (1) VMD decomposes the complex stock index price series into multiple simple intrinsic mode functions (IMFs) containing different frequency information; (2) For each IMF, PatchTST is used to learn its temporal patterns across different time scales, and the predictions from these patterns are aggregated additively to form the final forecast; (3) ASWL is introduced to capture the scale information of each subsequence, optimizing the resource allocation during model training. The main contributions of the paper are as follows:

(1) This study proposes a composite forecasting framework: VMD+PatchTST with ASWL. The VMD+PatchTST framework demonstrates robust predictive performance, addressing the limitations of previous methods in effectively learning temporal modes. ASWL further optimizes the average loss function of deep models in multivariate time series forecasting, creating a scale-weighted decomposition and aggregation framework. We applied the proposed model to multiple stock index datasets. Experimental results indicate that the VMD+PatchTST with ASWL framework provides more accurate predictions compared to previous models. 

(2) The superiority of the VMD+PatchTST with ASWL framework is demonstrated through various experiments, including direct forecasting methods with individual models, composite methods combining VMD with state-of-the-art deep models, and different composite methods with scale-weighted. Experimental results show that VMD+PatchTST with ASWL outperforms other methods in model evaluation metrics.

(3) This study innovatively proposes the adaptive scale-weighted layer. Without altering the aggregation method of sub-sequences, ASWL incorporates the scale information of the original sub-sequences, preventing inefficient computational resource allocation during model training. The validity of ASWL has been verified across multiple deep models. Experimental results demonstrate that the inclusion of ASWL significantly reduces the prediction error of composite methods.

(4) The VMD+PatchTST with ASWL composite forecasting framework performs well on the SP500, DJI, SSEC, and FTSE datasets. The proposed model achieves mean squared error (MSE) values of 7.69, 51.67, 13.29, and 19.91, and symmetric mean absolute percentage error (sMAPE) values of 0.42\%, 0.24\%, 0.46\%, and 0.29\%, respectively. Overall, the proposed VMD+PatchTST with ASWL framework demonstrates superior predictive accuracy and strong generalization capability compared to previous models.

The rest of the paper is structured as follows. The proposed VMD+PatchTST with ASWL framework is elaborated in Section~\ref{sec:methodology}. Section~\ref{sec:exp_set} details the experimental setup. 
The results are presented and analyzed in Section~\ref{sec:exp}. Finally, Section~\ref{sec:conclusion} concludes the paper.

\section{Methodology}
\label{sec:methodology}
\begin{figure*}[htbp]
	\centering
	\includegraphics[width=\textwidth]{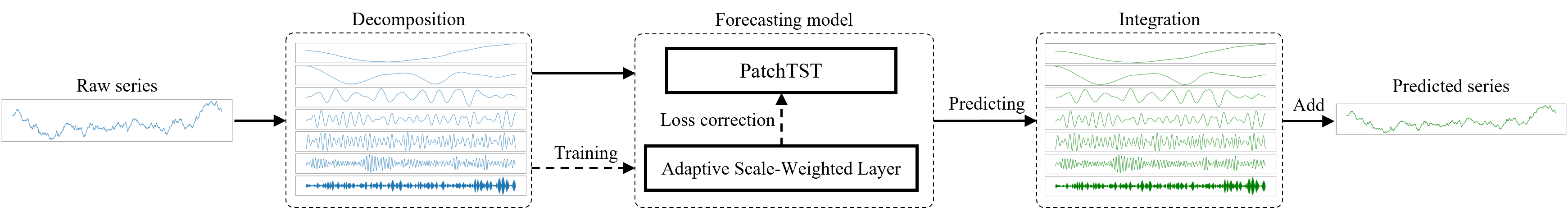}
	\caption{The flowchart of the proposed framework}
	\label{fig:flowchart}	
\end{figure*}

In this section, we will introduce the proposed decomposition-integrated framework, VMD+PatchTST with ASWL. As shown in Fig.~\ref{fig:flowchart}, this framework is composed of three parts: the decomposition module VMD, the forecasting model PatchTST, and the newly proposed adaptive scale-weighted layer. First, the raw series is decomposed into multiple subsequences by the decomposition module. Next, the forecasting model learns from the training dataset, with the adaptive scale-weighted layer responsible for loss correction during the training process. Finally, the forecasting model processes the test dataset to produce predicted subsequences, which are then aggregated in the integration module to obtain the final predicted series. The following provides a detailed explanation of each component.

\subsection{Variational modal decomposition}
\label{subsec:vmd}

The variational mode decomposition (VMD) algorithm is a novel time-frequency analysis method \cite{dragomiretskiy2013variational}, which decomposes a multi-component time series into multiple single-component amplitude-modulated (AM) and frequency-modulated (FM) signals. Unlike traditional empirical mode decomposition (EMD), VMD circumvents endpoint effects and pseudo-component issues encountered during iterative processes. Additionally, it exhibits superior robustness on complex time series that are often nonlinear and non-stationary. Specifically, the VMD decomposition process is a variational optimization process that decomposes the original time series $S(t)$ into K bandwidth-constrained intrinsic mode functions $s_m(t)$ and their corresponding center frequencies $v_m(t), (m=1,2,\cdots,M)$. The constrained bandwidth is estimated through demodulation signal estimation using the $L^2$ norm gradient. The mathematical expression is as follows:

\begin{equation}
	\begin{aligned}
		\min_{\{s_m\}, \{v_m\}} \left\{ \sum_{m=1}^{M} \left\| \partial_t \left[ \left( \delta(t) + 	\frac{j}{\pi t} \right) * s(t) \right] e^{-j v_m t} \right\|_2^2 \right\} \\
		\quad \text{s.t.} \quad
		\sum_m s_m(t) = S(t)
		\label{eq:(1)}
	\end{aligned}
\end{equation}

where $\delta(t)$ denotes an impulse function, $\sum_m := \sum_{m=1}^M$ is interpreted as the aggregate of all sub-modes. \Cref{eq:(1)} represents a typical reconstruction constraint problem. In order to address this issue, a quadratic penalty term $\alpha$ and a Lagrange multiplier $\lambda$ are introduced, thereby transforming the problem into an unconstrained one.

\begin{equation}
	\begin{aligned}
		\mathcal{L}(s_m,v_m,\lambda) &= \alpha\sum_m\left\| \partial_t \left[ \left( \delta(t)  + 	\frac{j}{\pi t} \right) * s(t) \right] e^{-j v_m t} \right\|_2^2 \\
		& \quad + \left\| f(t)-\sum_m s_m(t) \right\|_2^2 \\
		& \quad + \left\langle \lambda(t), f(t)-\sum_m s_m(t) \right\rangle
	\end{aligned}
	\label{eq:(2)}
\end{equation}

The alternating direction method of multipliers (ADMM) was employed for the solution of \cref{eq:(2)}. In conclusion, the iterative application of \cref{eq:(3),eq:(4),eq:(5)} until the condition in \cref{eq:(6)} is met represents the final stage of the process, we obtained the final output $\widetilde{s_m^{n+1}(\omega)},\ c_m^{n+1},\ \text{and}\ \widehat{\lambda^{n+1}(c)}$ of the VMD algorithm.

\begin{align}
	\widetilde{s_m^{n+1}(v)} &= \frac{\widehat{f(v)} - \sum_{i \neq k}s_m^{n+1}(v)+\widehat{\lambda^n(v)}}{1+2\alpha(v-v_m)^2}
	\label{eq:(3)} \\
	v_m^{n+1} &= \frac{\int_{0}^{\infty}v\left|\widetilde{s_m^{n+1}(v)}\right|^2dv}
	{\int_{0}^{\infty}\left|\widetilde{s_m^{n+1}(v)}\right|^2dv}
	\label{eq:(4)} \\
	\widehat{\lambda^{n+1}(v)} &= \widehat{\lambda^n(v)}+\tau(\widehat{f(v)}-\sum_m\left|s_m^{n+1}(v) \right|)
	\label{eq:(5)}		
\end{align}

\begin{equation}
	\frac{\left\| \widetilde{s_m^{n+1}}-\widetilde{s_m^{n}} \right\|^2}{\left\| \widetilde{s_m^{n}}  \right\|^2}<\varepsilon
	\label{eq:(6)}
\end{equation}

where $\tau$ is a noise tolerance parameter of VMD and $\varepsilon$ is a given discrimination accuracy,

\subsection{Forecasting Model}
\label{subsec:forecastmodel}
Transformer-based models have been extensively employed in the domain of time series modeling. Financial data, which is a prototypical example of a nonlinear and non-stationary time series, has prompted researchers to employ such models to accurately predict it, with the aim of achieving outcomes such as risk diversification and excess returns. The Transformer is an encoder-decoder framework entirely based on attention mechanisms. The encoder comprises stacked attention mechanisms, feed-forward neural networks, and skip connections, and is employed to transform time series into high-dimensional latent representations. In contrast, the decoder incorporates a masking mechanism to prevent information leakage and combines the original sequence and the high-dimensional representations derived from the encoder to facilitate predictions about the target series. However, the computational complexity of the Transformer model is \(O(L^2)\) (where \(L\) is the length of the time series), which can result in excessive computational demands for long sequences. Furthermore, time series at different time steps demonstrate varying degrees of correlation, whereas the input data for Transformers are point-wise tokens, which only contain information from a single step. The PatchTST model offers an effective solution to the aforementioned issues.	

PatchTST is a transformer-based approach that contains two principal modules: patching and channel independence \cite{nie2023a}. Patching enables the model to handle tokens of varying time steps, aggregating these into subsequences that capture locality and global semantic information. Channel-independence implies that each input token contains complete information from a single time series. In multivariate time series forecasting, this independence emphasizes the significance of each feature, thereby reducing model overfitting. 

\begin{figure}[htbp]
	\centering
	\begin{subfigure}{0.5\textwidth}
		\centering
		\includegraphics[width=\linewidth]{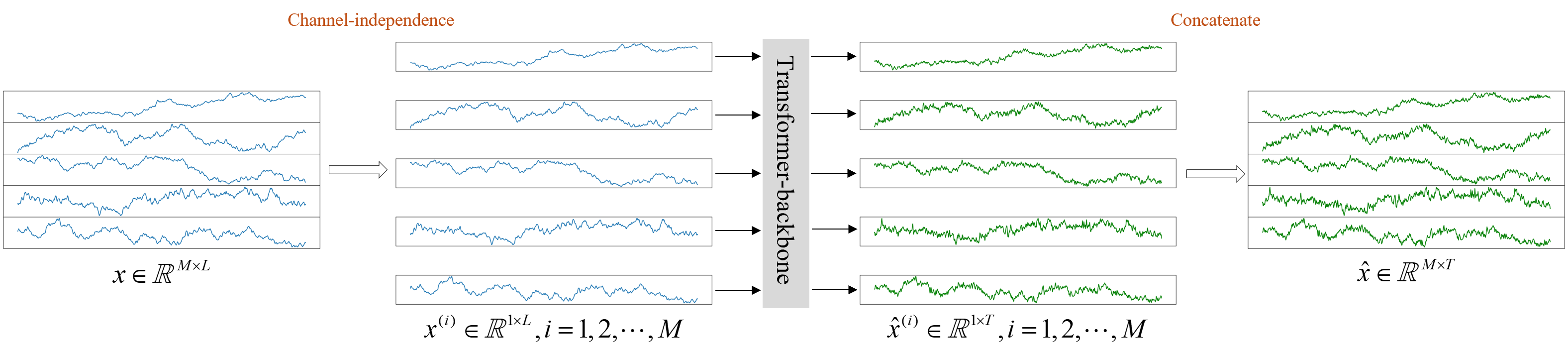} 
		\caption{PatchTST Overview}
		\label{fig1:subfig_a}
	\end{subfigure}\hspace{1cm}
	
	\begin{subfigure}{0.5\textwidth}
		\centering
		\includegraphics[width=\linewidth]{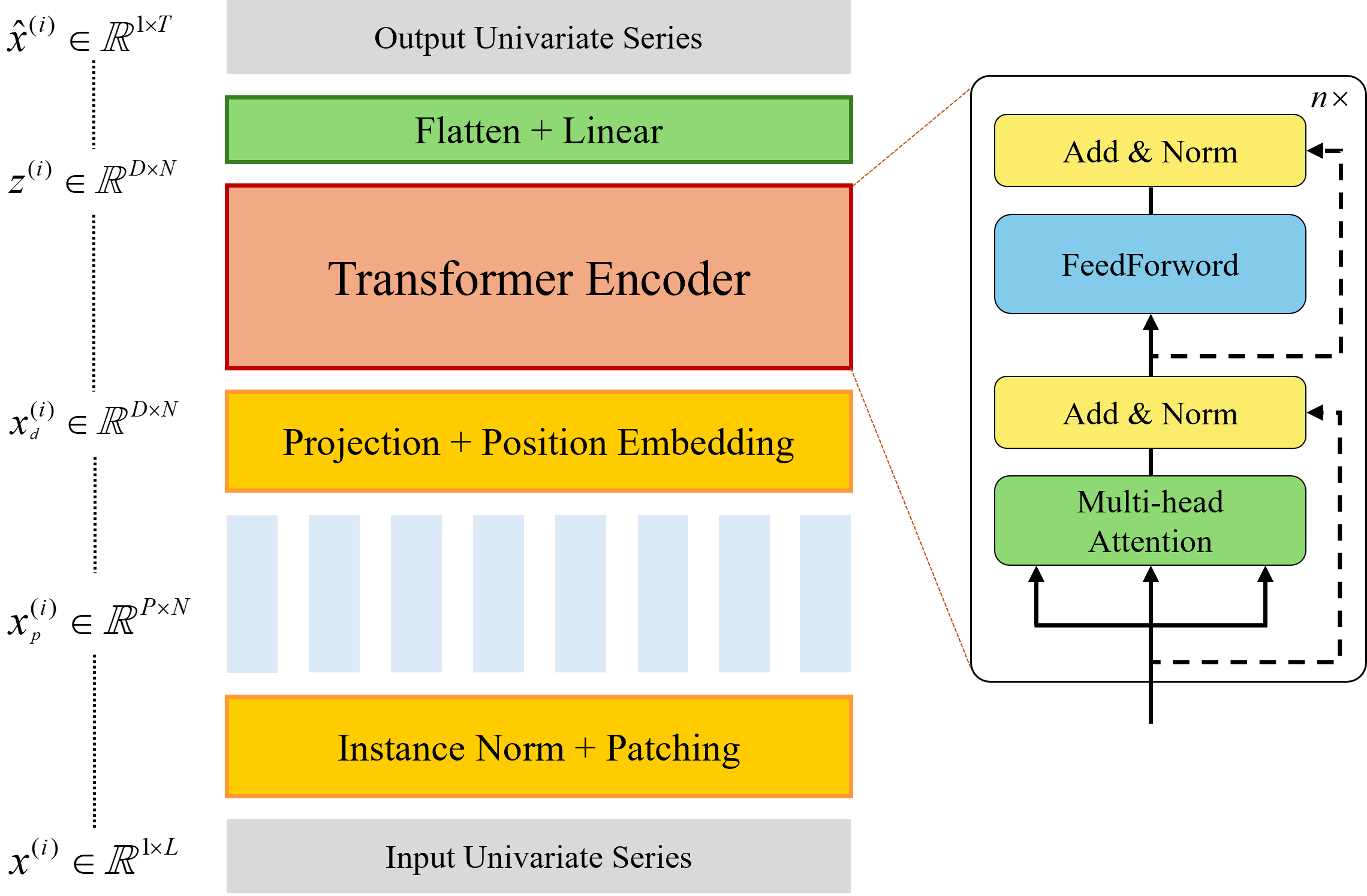} 
		\caption{Transformer Backbone}
		\label{fig1:subfig_b}
	\end{subfigure}
	\caption{The architecture of PatchTST}
	\label{fig1:main_fig}
\end{figure}

Fig.~\ref{fig1:main_fig} illustrates the architecture of PatchTST. The model consists of four major components: Forward Process, Patching, Transformer Encoder, and Instance Normalization. where each $x_t$ at time step t is a vector of dimension M. Given a set of multivariate time series, the current task is to forecast future values for a time horizon of $T: (\bm{x}_{L+1},\cdots,\bm{x}_{L+T})$ using a lookback window of $L:(\bm{x}_1,\cdots,\bm{x}_L)$ where each $x_t$ is an $m$-dimensional vector. 

\paragraph{(1) Forward Process} Assume the $i$-th univariate time series has a length of $L$ and starts from timestamp 1, denoted as $\bm{x}_{1:L}^{(i)}=(x_1^{(i)},x_2^{(i)},\cdots,x_L^{(i)})$, where \(i = 1, 2, \cdots, M\). The multivariate time series input is decomposed into M univariate time series \(\bm{x}^{(i)} \in \mathbb{R}^{1 \times L} \), which are each fed into the channel-independence Transformer backbone. Then the Transformer backbone will generate prediction results.
\begin{equation}
	\hat{\bm{x}}^{(i)}=\left(\hat{x}_{L+1}^{(i)},\hat{x}_{L+2}^{(i)},\cdots,\hat{x}_{L+T}^{(i)} \right)^T \in \mathbb{R}^{1 \times T}
	\label{eq:(7)}
\end{equation} 

\paragraph{(2) Patching} In this module, each input univariate time series is segmented into $N$ patches series $\bm{x}^{(i)}_p \in \mathbb{R}^{P \times N}$, where $N$ is given by $\lfloor \frac{L-P}{S} + 2 \rfloor$, with $P$ representing the length of each patch and $S$ denoting the stride. Notably, to preserve the information completeness of the patches sequence, the final value $x_L^{(i)} \in \mathbb{R}$ is repeated $S$ times to extend to the end of the original sequence before patching.

\paragraph{(3) Transformer Encoder} PatchTST uses a vanilla transformer encoder as the backbone model for extracting latent representations from input signals. The patches are mapped into a $D$-dimensional latent space by a learnable linear projection $ W_p \in \mathbb{R}^{D \times P}$. In addition, a learnable additive positional encoding $W_{pos} \in \mathbb{R}^{D \times N}$ is introduced to maintain temporal order consistency of patches:

\begin{equation}
	x_d^{(i)}=\mathbf{W}_p x_p^{(i)}+
	\mathbf{W}_{pos}
	\label{eq:(8)}		
\end{equation}

where $x_d^{(i)} \in \mathbb{R}^{D \times N}$ represents the input fed into the Transformer encoder.

Then, the generation of query, key, and value matrices will be performed by each head in the multi-head attention mechanism.

\begin{align}
	\bm{Q}_h^{(i)}=(x_d^{(i)})^T \mathbf{W}_h^Q \\
	\bm{K}_h^{(i)}=(x_d^{(i)})^T \mathbf{W}_h^K \\
	\bm{V}_h^{(i)}=(x_d^{(i)})^T \mathbf{W}_h^V
	\label{eq:(9-11)}		
\end{align}

The scaled dot product attention output $\mathbf{O}_h^{(i)} \in \mathbb{R}^{D \times N}$ is then calculated.

\begin{equation}
	\begin{aligned}
		(\mathbf{O}_h^{(i)})^T &= \text{Attention}\left(\bm{Q}_h^{(i)}, \bm{K}_h^{(i)}, \bm{V}_h^{(i)}\right) \\ 
		&= \text{Softmax} \left(\frac{Q_h^{(i)} K_h^{(i)^T}}{\sqrt{d_k}}V_h^{(i)}\right)
	\end{aligned}
	\label{eq:(12)}
\end{equation}

where $\mathbf{W}_h^Q, \mathbf{W}_h^K \in \mathbb{R}^{D \times R_k}$ and $\mathbf{W}_h^V \in \mathbb{R}^{D \times D}$.

Note that both BatchNorm layers and a feed forward network with residual connections are present within the multi-head attention block. The latent representations after the multi-head attention are denoted as $z^{(i)} \in \mathbb{R}^{D \times N}$. Finally, these representations are processed through a flatten layer with linear head to obtain the final prediction results, as shown in \cref{eq:(7)}.

\paragraph{(4) Instance Normalization} Ulyanov, D. et al. proposed a technique that effectively captures and processes the style information of each sample, particularly useful for image generation and style transfer tasks \cite{ulyanov2016instance}. This technique normalizes each sample independently, which improves the quality and consistency of the generated data while reducing batch size dependency. We perform instance normalization on each sequence $\bm{x}^{(i)}$ before patching and then add the mean and bias back to the output predictions.

\subsection{Adaptive Scale-Weighted Layer}
\label{subsec:ASWL}
\begin{figure}[htbp]
	\centering
	\includegraphics[width=0.5\textwidth]{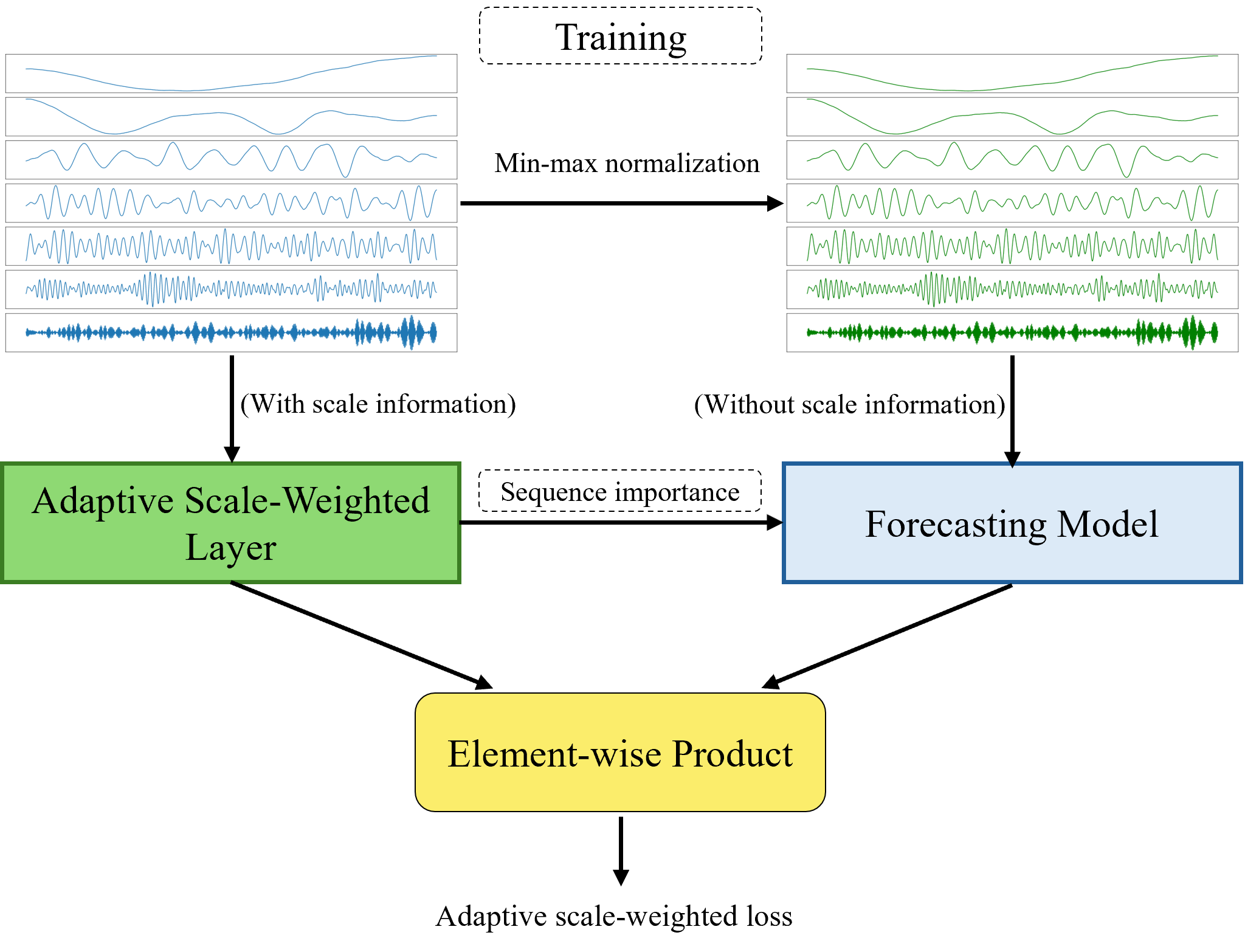}
	\caption{The Adaptive scale-weighted layer}
	\label{fig:ASWL}
\end{figure}

The raw time series $S(t)$ is decomposed by VMD into several IMFs \( \bm{X}_{1:L}=(\bm{x}^{(1)}_{1:L},\bm{x}^{(2)}_{1:L},\cdots,\bm{x}^{(M)}_{1:L}) \). Each IMF $ \bm{X}^{(i)}_{1:L} $ contains multi-scale features. The forecasting model then learns from each IMF to produce the forecast outputs $\hat{\bm{x}}_{L+1:L+T}=(\hat{\bm{x}}^{(1)}_{L+1:L+T},\hat{\bm{x}}^{(2)}_{L+1:L+T},\cdots,\hat{\bm{x}}^{(M)}_{L+1:L+T})$. The final prediction is obtained by simply summing these outputs  $ \hat{\bm{x}} = \sum_{1}^{M} \hat{\bm{x}}_{L:T}$. Due to the inherent sensitivity of deep models to the scale of the data, we normalize the data to the 0-1 range (see \cref{eq:(13)}), which means a lack of scale information. Furthermore, for multivariate time series problems, the total loss function during training is generally the simple sum of the losses from each sequence, meaning that the model treats all subsequences with equal importance. In reality, these IMFs possess different frequency and scale information. To integrate this information effectively, we introduce an adaptive scale-weighted layer (ASWL) into the model, as shown in Fig.~\ref{fig:ASWL}.

\begin{equation}
	x_{0-1}=\frac{x-x_{min}}{x_{max}-x_{min}}
	\label{eq:(13)}
\end{equation}

The design of the ASWL aims to dynamically adjust the weights according to the importance of each sub-sequence and its contribution to the overall prediction. Specifically, it consists of a linear layer without bias $ \bm{W}_M^{ASWL}=(w^{(1)},w^{(2)}, \cdots,w^{(M)}) $, which integrates the multi-scale information of the sequences and assigns different weights to the loss of high and low frequency components during training. As a result, the model's ability to predict complex time series is enhanced, along with improved robustness and generalization performance. Finally, the prediction results are obtained by replacing simple addition with an element-wise product.

\begin{equation}
	\scalebox{1}{
	$\text{Adaptive scale-weighted loss}= \sum_{i=1}^{M}\text{Loss}(\bm{x}_{L+1:L+T}^{(i)}-\hat{\bm{x}}_{L+1:L+T}^{(i)}) w^{(i)}
	$}
\end{equation}

\section{Experimental Setup}
\label{sec:exp_set}

\subsection{Data}
To validate the performance of the proposed method, we use daily closing prices of global stock indices as experimental data, which can be obtained from the Wind database. Specifically, we utilize four stock indices: the Standard \& Poor's 500 Index (SP500), the Dow Jones Industrial Average (DJI), the Shanghai Composite Index (SSEC), and the Financial Times Stock Exchange 100 Index (FTSE), covering the period from January 2000 to June 2024. After excluding unavailable data, the number of observations for each index is 6,123, 6,145, 5,920, and 5,893, respectively.

\begin{figure}[htbp]
	\centering
	\includegraphics[width=0.5\textwidth]{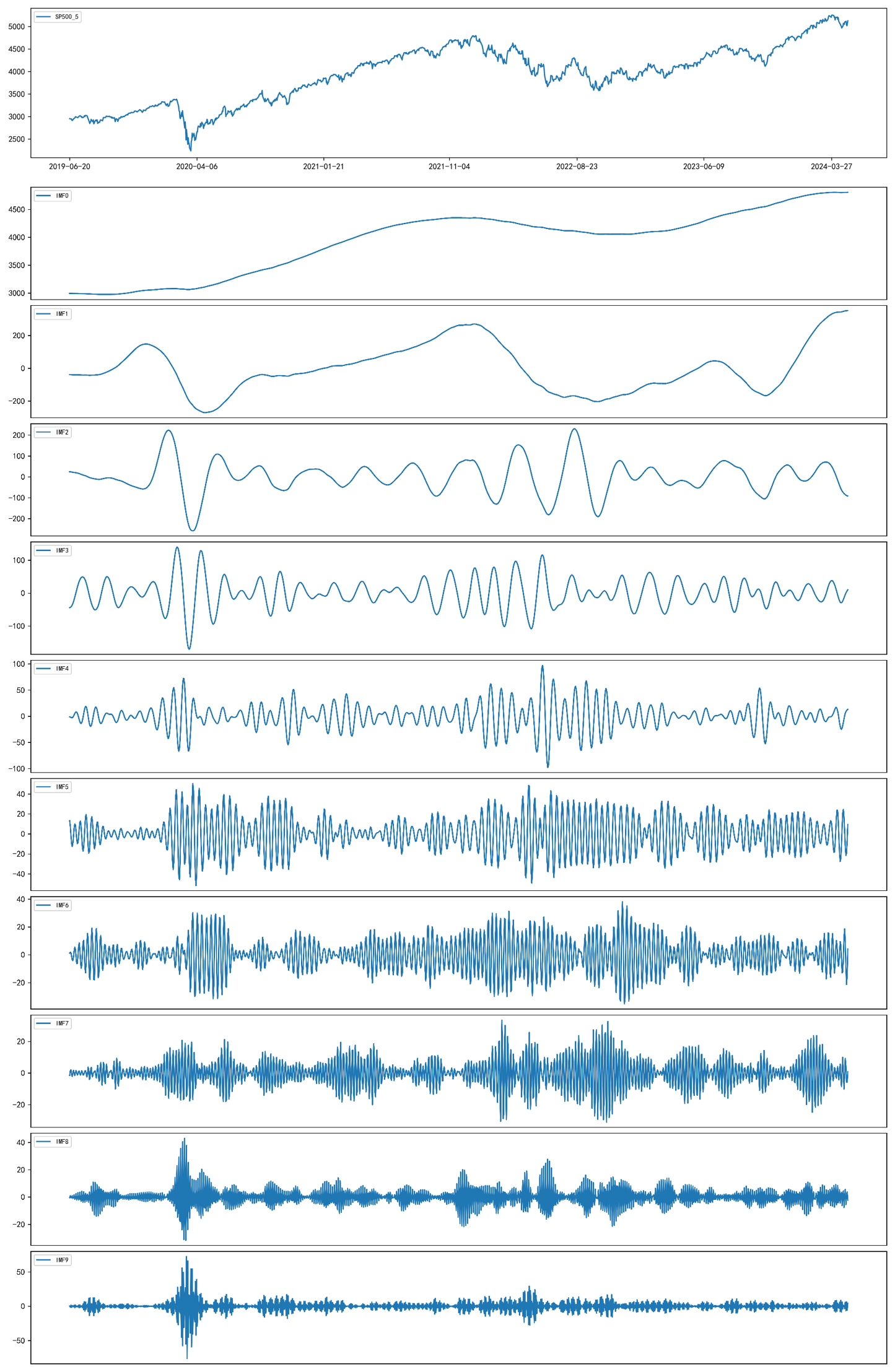}
	\caption{The decomposed components of the fifth period SP500 by VMD}
	\label{fig:vmd}
\end{figure}

Since the characteristics of these indices are time-varying, we divide the dataset into five equal parts, with 80\% of each part used as the training set and 20\% as the test set. Figure~\ref{fig:vmd} shows the prices of the SP500 index in the fifth period and the corresponding VMD decomposition components. It can be observed that the earlier IMFs have a larger range, reflecting the low-frequency characteristics of the original sequence, while the later IMFs have a smaller range, reflecting the high-frequency characteristics of the original sequence.

\subsection{Evaluate criteria}
To comprehensively evaluate the model's prediction results, we use mean squared error (MSE) and symmetric mean absolute percentage error (sMAPE) as the evaluation metrics of experiments. The formulas are as follows:

\begin{align}
	MSE = \frac{1}{MT}\sum_{i=1}^{M} \sum_{t=L+1}^{L+T}(x^{(i)}_t-\hat{x}^{(i)}_t)^2 \\
	sMAPE = \frac{2}{MT}\sum_{i=1}^{M} \sum_{t=L+1}^{L+T} \frac{\left| x_t^{(i)} - \hat(x^{(i)}_t) \right|}{ \left| x^{(i)}_t \right| - \left| \hat{x}^{(i)}_t \right|}
\end{align}
where $x_j^{(i)}$ and $\hat{x}_j^{(i)}$ represent the actual value and the predicted value of the $i$-th sequence at time $t$, respectively. $T$ denotes the number of test data, and $M$ denotes the number of sequences in the testset.

\subsection{Comparable models and parameter setting}
To evaluate the performance of our proposed prediction model, we select several state-of-the-art deep models as baselines: CNN-LSTM\cite{kim2019predicting}, Informer\cite{zhou2021informer}, Autoformer\cite{wu2021autoformer}, Non-stationary Transformer\cite{liu2022non}, and traditional models: Prophet\cite{taylor2018forecasting}, ARIMA\cite{box1976time}. First, we directly compare these individual forecasting models, which are well-recognized for time series forecasting. Next, we combine VMD with deep forecasting models to predict the four stock price datasets, demonstrating the improvement in financial time series forecasting performance. Finally, we introduce ASWL and compare the performance of each model. The parameter settings for the models are shown in Table~\ref{table:para_set}. All experiments were conducted using Python 3.11 and PyTorch 2.1. The computational tasks were accelerated with an NVIDIA GeForce RTX 4060 graphics card.

\begin{table}[ht]
	\centering
	\caption{The hyperparameters of the models}
	\begin{tabular}{ccc}
		\toprule
		Method & Description & Value \\ 
		\midrule
		VMD & Number of modes & 10 \\ 
		\midrule
		\multirow{8}{*}{\begin{tabular}[c]{@{}c@{}}Transformer-based models\\ \vspace{0.3cm} \\ LSTM\end{tabular}} 
		& Dimension model & 512 \\  
		& Encoder Layers & 2 \\  
		& Decoder Layers & 1 \\  
		& Learning Rate & 0.001 \\
		& Batch size & 32 \\  
		& Iterations & 5 \\ 
		& Epochs & 150 \\ 
		& Optimizer & Adam \\ 
		\bottomrule
	\end{tabular}
	\label{table:para_set}
\end{table}

\section{Results and analysis}
\label{sec:exp}
\subsection{Single model prediction performance}
\label{Single model prediction performance}
In this subsection, the prediction results of PatchTST for four stock index prices are compared in terms of MSE and sMAPE with those of other forecasting models. Table~\ref{table:single_model} presents the forecasting performance of each individual model. Compared to other models—Informer, Autoformer, Non-stationary Transformer, Prophet, and ARIMA—PatchTST has the second-lowest error values across all four datasets, only surpassed by the Non-stationary Transformer. This indicates that these two models offer optimal fitting performance. The sMAPE values for these models are on the SP500 dataset as follows: Informer at 12.68\%, Autoformer at 3.68\%, Non-stationary Transformer at 1.12\%, PatchTST at 1.38\%, CNN-LSTM at 5.69\%, Prophet at 8.9\% and ARIMA at 40.38\%. ARIMA, a traditional model, demonstrates poor performance in learning the characteristics of non-stationary and nonlinear stock time series, leading to its exclusion from subsequent experiments. Although CNN-LSTM performs worse than Autoformer and Non-stationary Transformer, it remains a viable model choice. Transformer-based models generally exhibit strong predictive capabilities. Among these, the De-stationary Attention mechanism in the Non-stationary Transformer and the patching approach in PatchTST are likely key factors enabling these models to effectively handle complex non-stationary time series. However, direct modeling of sequences still faces significant challenges in terms of accuracy.

\begin{table*}[htb]
	\centering
	\caption{Forecasting results of different individual models on four stock index datasets}
	\begin{tabular}{cccccccccl}
		\toprule
		\multicolumn{2}{c}{Model} &
		\multicolumn{2}{c}{Informer} &
		\multicolumn{2}{c}{Autoformer} &
		\multicolumn{2}{c}{\begin{tabular}[c]{@{}c@{}}Non-stationary \\ Transformer\end{tabular}} &
		\multicolumn{2}{c}{PatchTST} \\ \midrule
		Data       & Metric       & MSE      & sMAPE  & MSE        & sMAPE  & MSE      & sMAPE  & MSE                  & \multicolumn{1}{c}{sMAPE}  \\
		\midrule
		\multicolumn{2}{c}{SP500} & 226.2198 & 0.1268 & 83.4173    & 0.0368 & 25.75    & 0.0112 & 31.5375              & \multicolumn{1}{c}{0.0138} \\
		\multicolumn{2}{c}{DJI}   & 1615.535 & 0.1017 & 1072.5267 & 0.0545 & 189.6984 & 0.0086 & 255.4302             & \multicolumn{1}{c}{0.0117} \\
		\multicolumn{2}{c}{SSEC}  & 57.23341 & 0.0192 & 137.7081   & 0.047  & 41.4493  & 0.0128 & 47.5158              & \multicolumn{1}{c}{0.015}  \\
		\multicolumn{2}{c}{FTSE}  & 300.0001 & 0.0441 & 220.6956   & 0.0305 & 52.5006  & 0.0065 & 64.8397              & \multicolumn{1}{c}{0.0082} \\ \midrule
		\multicolumn{2}{c}{Model} &
		\multicolumn{2}{c}{CNN-LSTM} &
		\multicolumn{2}{c}{Prophet} &
		\multicolumn{2}{c}{ARIMA} &
		\multicolumn{2}{c}{} \\ 
		\midrule
		Data       & Metric       & MSE      & sMAPE  & MSE        & sMAPE  & MSE      & sMAPE  & \multicolumn{1}{l}{} &                            \\
		\midrule
		\multicolumn{2}{c}{SP500} & 129.2682 & 0.0569 & 229.6064   & 0.089  & 830.758  & 0.4038 & \multicolumn{1}{l}{} &                            \\
		\multicolumn{2}{c}{DJI}   & 871.9298 & 0.0458 & 1255.266   & 0.0788 & 1903.129 & 0.1218 & \multicolumn{1}{l}{} &                            \\
		\multicolumn{2}{c}{SSEC}  & 75.0816  & 0.022  & 725.988    & 0.3519 & 736.1652 & 0.347  & \multicolumn{1}{l}{} &                            \\
		\multicolumn{2}{c}{FTSE}  & 80.7218  & 0.0105 & 764.029    & 0.1461 & 396.4683 & 0.061  & \multicolumn{1}{l}{} &                            \\ \bottomrule
	\end{tabular}
	\label{table:single_model}
\end{table*}

\subsection{Performance of VMD-augmented deep models}
In light of the findings presented in Section~\ref{Single model prediction performance} regarding individual forecasting models, it can be concluded that the Non-stationary Transformer and PatchTST models demonstrate the most effective predicted performance. This section discusses the performance of VMD-augmented deep models. The table presents the predicted performance of VMD in conjunction with a variety of deep models across four stock indices datasets. First, A comparison of Tables~\ref{table:single_model}~and~\ref{table:VMD+models} reveals that the MSE and sMAPE of the VMD+deep models framework are markedly lower than those of the individual forecasting models. This suggests that VMD can effectively decompose complex time series into sub-series containing different frequency information for model learning, and this approach is superior to direct learning of the series. Second, the results presented in Table~\ref{table:VMD+models} indicate that, among the different models, the VMD+PatchTST exhibits the lowest MSE and sMAPE across all stock indices datasets, with the exception of the MSE associated with the SP500. For instance, on the DJI dataset, the VMD+PatchTST model attains an MSE of 67.66 and an sMAPE of 0.35\%, outperforming other models in prediction accuracy. Similarly, it also demonstrates superior performance on the SP500, SSEC, and FTSE datasets, thereby providing evidence of its effectiveness in capturing complex time patterns and providing accurate forecasts. In contrast, although the VMD+CNN-LSTM and VMD+Autoformer models demonstrate comparable performance, their relatively higher MSE and sMAPE values indicate that they are less effective than VMD+PatchTST in capturing data patterns. Finally, the superior predicted performance of PatchTST over the Non-stationary Transformer through VMD augmentation further supports the conclusion that PatchTST is the optimal deep model for modelling multi-frequency scale information.

\begin{table*}[htb]
	\centering
	\caption{Forecasting results of VMD+deep models on four stock index datasets}
	\resizebox{\textwidth}{!}{
		\begin{tabular}{lccccccccccc}
			\toprule
			\multicolumn{2}{c}{Model} &
			\multicolumn{2}{c}{VMD+Informer} &
			\multicolumn{2}{c}{VMD+Autoformer} &
			\multicolumn{2}{c}{\begin{tabular}[c]{@{}c@{}}VMD+Non-stationary \\ Transformer\end{tabular}} &
			\multicolumn{2}{c}{VMD+PatchTST} &
			\multicolumn{2}{c}{VMD+CNN-LSTM} \\
			\midrule
			Data & Metric & \multicolumn{1}{c}{MSE} & \multicolumn{1}{c}{sMAPE} & \multicolumn{1}{c}{MSE} & \multicolumn{1}{c}{sMAPE} & \multicolumn{1}{c}{MSE} & \multicolumn{1}{c}{sMAPE} & \multicolumn{1}{c}{MSE} & \multicolumn{1}{c}{sMAPE} & \multicolumn{1}{c}{MSE} & \multicolumn{1}{c}{sMAPE} \\
			\midrule
			\multicolumn{2}{c}{SP500} & 109.4633 & 0.0707 & 92.9934 & 0.042 & 13.3046 & 0.0062 & 13.3369 & 0.0057 & 190.316 & 0.0933 \\
			\multicolumn{2}{c}{DJI}    & 924.4736 & 0.0547 & 563.0211 & 0.0282 & 105.3541 & 0.0051 & 67.66135 & 0.0035 & 1328.4645 & 0.0702 \\
			\multicolumn{2}{c}{SSEC}   & 154.3022 & 0.0479 & 101.7248 & 0.0356 & 26.7517 & 0.0082 & 17.8474 & 0.006 & 200.6226 & 0.0514 \\
			\multicolumn{2}{c}{FTSE}   & 158.2544 & 0.0279 & 134.3754 & 0.0183 & 26.5517 & 0.0034 & 23.1255 & 0.0031 & 145.233 & 0.0196 \\
			\bottomrule
		\end{tabular}
	}
	\label{table:VMD+models}
\end{table*}

Next, we analyze the learning capability of VMD+Models for each IMF. The forecasting results in Table~\ref{table:IMF10} highlight the performance of different models across various stock indices datasets, which are decomposed into intrinsic mode functions from IMF0 (low frequency) to IMF9 (high frequency). The VMD+PatchTST model stands out as the most exceptional, followed by the VMD+Non-stationary Transformer, both exhibiting lower MSE and sMAPE values. Notably, for low-frequency components (IMF0), the VMD+Informer model demonstrates robust performance in capturing long-term trends, particularly in the DJI dataset, with relatively low MSE and sMAPE values. However, this model's performance significantly declines when handling high-frequency IMFs, indicating its limitations in addressing short-term fluctuations and noise.

\begin{figure*}[ht]
	\centering
	\includegraphics[width=0.6\textwidth]{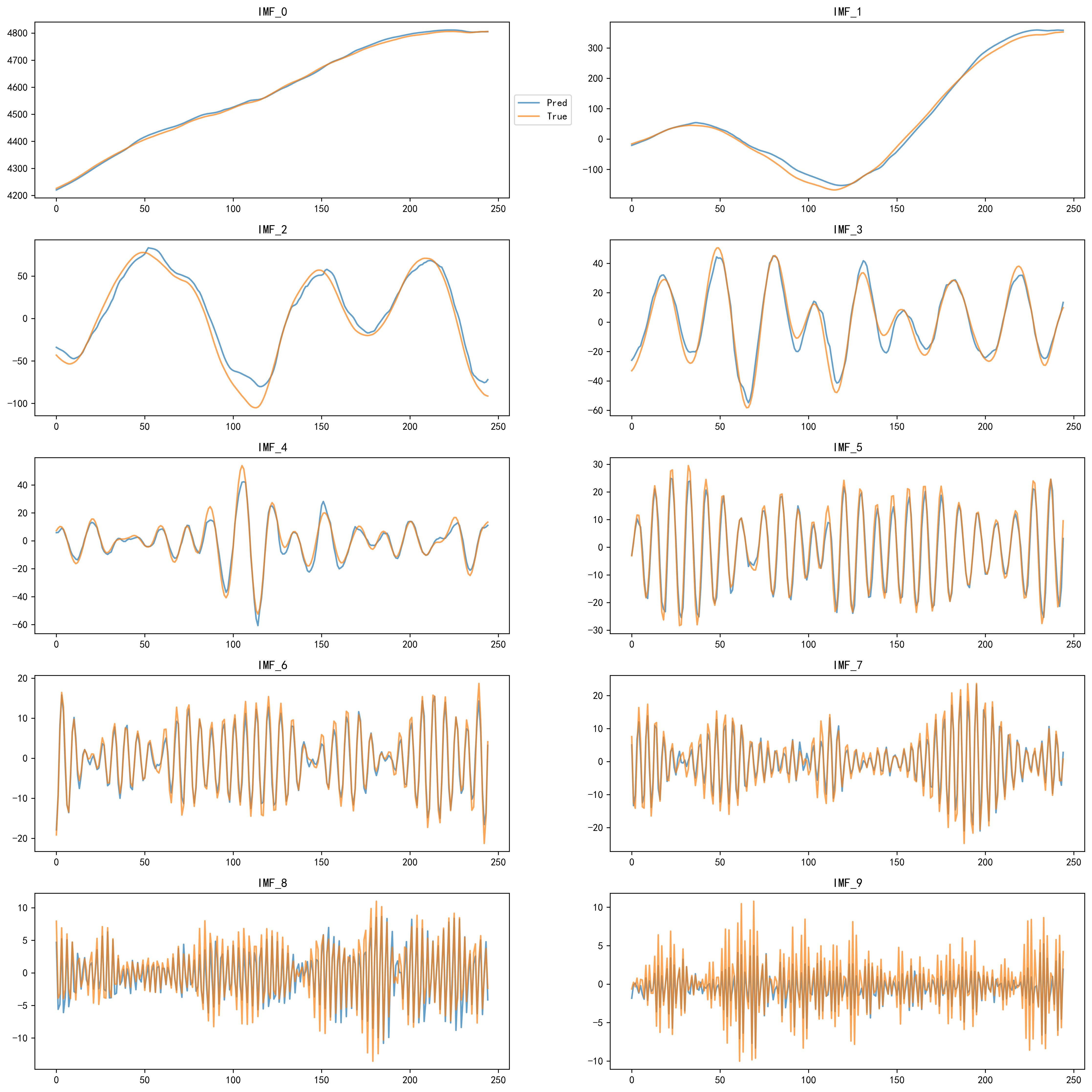}
	\caption{Forecasting results of IMFs for the fifth period SP500 using VMD+PatchTST}
	\label{fig:imf_pred}
\end{figure*}

In contrast, the VMD+PatchTST model demonstrates superior performance in forecasting high-frequency components (IMF9), achieving the lowest MSE and sMAPE values across multiple datasets. The model's ability to effectively capture and predict short-term fluctuations and high-frequency noise establishes it as a strong tool for managing high-frequency volatility in financial time series forecasting. The VMD+CNN-LSTM model consistently underperforms in predicting low-frequency components, indicating potential deficiencies in capturing broader market trends and long-term dependencies. To further illustrate the performance of the VMD+PatchTST model, Figure~\ref{fig:imf_pred} presents a predictive plot for the fifth period SP500 dataset. This visualization highlights the model's accuracy in forecasting and capturing the underlying patterns of the time series, underscoring its reliability. These results emphasize the importance of model selection and the crucial role of frequency decomposition techniques like VMD in improving the accuracy of non-stationary financial time series forecasting.

\begin{table*}[ht]
	\centering
	\caption{Forecasting results of VMD+deep models on IMFs of four stock index datasets}
	\resizebox{\textwidth}{!}{
		\begin{tabular}{lccccccccccc}
			\toprule
			\multicolumn{2}{c}{Model} &
			\multicolumn{2}{c}{VMD+Informer} &
			\multicolumn{2}{c}{VMD+Autoformer} &
			\multicolumn{2}{c}{\begin{tabular}[c]{@{}c@{}}VMD+Non-stationary \\ Transformer\end{tabular}} &
			\multicolumn{2}{c}{VMD+PatchTST} &
			\multicolumn{2}{c}{VMD+CNN-LSTM} \\
			\midrule
			Data & Metric & \multicolumn{1}{c}{MSE} & \multicolumn{1}{c}{sMAPE} & \multicolumn{1}{c}{MSE} & \multicolumn{1}{c}{sMAPE} & \multicolumn{1}{c}{MSE} & \multicolumn{1}{c}{sMAPE} & \multicolumn{1}{c}{MSE} & \multicolumn{1}{c}{sMAPE} & \multicolumn{1}{c}{MSE} & \multicolumn{1}{c}{sMAPE} \\
			\midrule
			\multirow{10}{*}{SP500} & IMF0   & 104.6862 & 0.0594 & 48.3218 & 0.0261 & 4.4391 & 0.0018 & 4.1730 & 0.0022 & 169.0538 & 0.0801 \\
			& IMF1   & 34.6890 & 0.7262 & 53.4712 & 1.1588 & 8.7961 & 0.2734 & 7.5329 & 0.2647 & 39.5661 & 0.7662 \\
			& IMF2   & 9.3071  & 0.6228 & 20.3370 & 0.9984 & 4.7909 & 0.3240 & 5.0623 & 0.3187 & 12.0688 & 0.6654 \\
			& IMF3   & 6.1501  & 0.7154 & 8.9802  & 1.1018 & 2.9863 & 0.3885 & 2.6611 & 0.3489 & 7.5184 & 0.7855 \\
			& IMF4   & 3.9779  & 0.8386 & 4.0501  & 0.8910 & 2.0078 & 0.4418 & 1.7083 & 0.4186 & 5.4847 & 0.9436 \\
			& IMF5   & 3.5378  & 0.8717 & 3.4724  & 0.8047 & 1.8253 & 0.4782 & 1.4530 & 0.4147 & 3.7186 & 0.8191 \\
			& IMF6   & 2.3958  & 0.9989 & 2.9184  & 1.0687 & 1.4886 & 0.5899 & 1.0515 & 0.4798 & 2.5440 & 0.9623 \\
			& IMF7   & 2.2047  & 1.0199 & 3.1126  & 1.1466 & 1.1897 & 0.5664 & 1.0756 & 0.5347 & 2.3122 & 0.9517 \\
			& IMF8   & 2.0011  & 1.1162 & 2.1320  & 1.0507 & 0.8218 & 0.5016 & 0.7784 & 0.4777 & 1.9916 & 0.9400 \\
			& IMF9   & 2.2307  & 0.9857 & 3.2663  & 1.1135 & 0.7287 & 0.4854 & 0.8705 & 0.4567 & 2.0858 & 1.0449 \\
			\midrule
			\multirow{10}{*}{DJI} & IMF0   & 881.2494 & 0.0518 & 414.7977 & 0.0212 & 41.2235 & 0.0026 & 22.5690 & 0.0015 & 1322.3190 & 0.0709 \\
			& IMF1   & 101.9526 & 0.5656 & 247.7687 & 0.8887 & 65.1011 & 0.2270 & 37.0432 & 0.1967 & 140.5444 & 0.4859 \\
			& IMF2   & 48.8912 & 0.4939 & 122.4842 & 0.8060 & 38.9738 & 0.3527 & 26.1881 & 0.2268 & 95.6606 & 0.6242 \\
			& IMF3   & 52.9501 & 0.8148 & 88.6666 & 0.9732 & 24.0203 & 0.4071 & 16.0602 & 0.3136 & 71.3888 & 0.7990 \\
			& IMF4   & 34.0636 & 0.8136 & 46.1487 & 0.9099 & 15.6798 & 0.4387 & 13.1548 & 0.3655 & 42.3952 & 0.8097 \\
			& IMF5   & 25.8800 & 0.8597 & 43.9634 & 1.2727 & 11.8995 & 0.5070 & 10.0309 & 0.4672 & 25.8331 & 0.8780 \\
			& IMF6   & 18.7336 & 0.9504 & 25.4960 & 1.1284 & 9.2516 & 0.5422 & 8.8975 & 0.5266 & 23.4215 & 0.9829 \\
			& IMF7   & 18.5252 & 1.1822 & 22.2244 & 1.1467 & 7.1379 & 0.5222 & 6.5823 & 0.5119 & 18.4661 & 1.0438 \\
			& IMF8   & 15.3304 & 1.0190 & 19.5763 & 1.0812 & 6.0784 & 0.4701 & 6.1080 & 0.4331 & 16.3602 & 0.9258 \\
			& IMF9   & 14.0704 & 0.9816 & 20.8751 & 1.1136 & 5.3314 & 0.4756 & 6.4820 & 0.4870 & 16.6410 & 0.9576 \\
			\midrule
			\multirow{10}{*}{SSEC} & IMF0 & 90.6425 & 0.0300 & 81.5545 & 0.0281 & 7.0418 & 0.0022 & 2.8196 & 0.0009 & 133.3294 & 0.0365 \\
			& IMF1 & 68.0821 & 0.5650 & 118.5690 & 0.8114 & 20.8979 & 0.1510 & 8.2516 & 0.1034 & 97.0275 & 0.5563 \\
			& IMF2 & 14.1476 & 0.5112 & 28.2947 & 0.9331 & 9.3299 & 0.3266 & 8.3446 & 0.3408 & 14.6809 & 0.5560 \\
			& IMF3 & 8.5492 & 0.5725 & 16.6036 & 0.9354 & 6.2977 & 0.4140 & 5.7454 & 0.3846 & 9.3226 & 0.5454 \\
			& IMF4 & 6.3716 & 0.7025 & 8.2352 & 0.8494 & 4.3084 & 0.4749 & 3.5381 & 0.4131 & 7.1194 & 0.7342 \\
			& IMF5 & 3.9446 & 0.7643 & 5.4684 & 0.8432 & 3.0392 & 0.5495 & 2.3897 & 0.4722 & 4.7530 & 0.7969 \\
			& IMF6 & 4.6305 & 1.1629 & 6.8741 & 1.1762 & 3.1819 & 0.7013 & 1.9989 & 0.5512 & 3.8091 & 0.9524 \\
			& IMF7 & 2.3875 & 0.7614 & 3.9233 & 1.0765 & 2.2947 & 0.6143 & 1.5499 & 0.5133 & 2.2586 & 0.8112 \\
			& IMF8 & 2.4269 & 0.8515 & 3.3819 & 0.9478 & 1.9860 & 0.5727 & 1.2322 & 0.4530 & 2.7392 & 0.9881 \\
			& IMF9 & 2.1136 & 0.8922 & 3.0212 & 1.0822 & 1.3233 & 0.5500 & 0.9857 & 0.4445 & 1.6832 & 0.8293 \\
			\midrule
			\multirow{10}{*}{FTSE} & IMF0 & 143.9905 & 0.0265 & 98.0628 & 0.015 & 9.0894 & 0.0013 & 5.974 & 0.0009 & 128.5021 & 0.0181 \\
			& IMF1 & 38.6864 & 0.6948 & 73.9767 & 0.9996 & 14.5133 & 0.265 & 13.2204 & 0.2249 & 45.9173 & 0.7286 \\
			& IMF2 & 18.377 & 0.7308 & 33.1269 & 0.9643 & 11.5225 & 0.3454 & 7.4078 & 0.2765 & 16.3603 & 0.5186 \\
			& IMF3 & 12.8989 & 0.6977 & 20.275 & 0.906 & 8.2836 & 0.4459 & 4.8054 & 0.3445 & 11.0386 & 0.5583 \\
			& IMF4 & 11.874 & 0.9704 & 12.3807 & 0.8382 & 5.2706 & 0.4664 & 3.7333 & 0.3463 & 9.2143 & 0.635 \\
			& IMF5 & 10.7521 & 1.0097 & 10.4225 & 1.0195 & 3.8819 & 0.5054 & 3.3823 & 0.4584 & 9.4255 & 0.8191 \\
			& IMF6 & 5.6245 & 0.9335 & 6.3888 & 0.9777 & 3.288 & 0.5596 & 2.9862 & 0.5579 & 6.1565 & 0.8159 \\
			& IMF7 & 7.6979 & 1.1752 & 6.2377 & 1.0806 & 2.6668 & 0.567 & 2.4063 & 0.5408 & 5.9534 & 1.0595 \\
			& IMF8 & 4.2372 & 1.0389 & 5.4839 & 1.0774 & 1.8396 & 0.5092 & 2.0293 & 0.4915 & 4.6951 & 0.9282 \\
			& IMF9 & 3.9175 & 1.0182 & 5.5418 & 1.2301 & 1.4642 & 0.4695 & 1.4324 & 0.4121 & 3.9417 & 1.0133 \\
			\bottomrule
		\end{tabular}
	}
	\label{table:IMF10}
\end{table*}

\subsection{Enhanced prediction with VMD, ASWL, and deep models}

It is well-established that IMFs represent the decomposed sequences of stock indices from low-frequency to high-frequency components (see Fig.~\ref{fig:vmd}), and therefore, their scales decrease progressively. This change in scale is also reflected in the MSE and sMAPE values (see Tables~\ref{table:VMD+models+ASWL} and \ref{table:IMF10_ASWL}). This observation indicates that the VMD-augmented models do not account for the original scale information of the IMFs. Consequently, we introduced the ASWL module to incorporate this information during model training. Tables X and Y respectively present the forecasting results of VMD+models with ASWL on four stock index datasets and their IMFs.

\begin{table*}[htbp]
	\centering
	\caption{Forecasting results of VMD+deep models with ASWL on four stock index datasets}
	\resizebox{\textwidth}{!}{
		\begin{tabular}{lccccccccc}
			\toprule
			\multicolumn{2}{c}{Model with ASWL} &
			\multicolumn{2}{c}{VMD+Informer} &
			\multicolumn{2}{c}{VMD+Autoformer} &
			\multicolumn{2}{c}{\begin{tabular}[c]{@{}c@{}}VMD+Non-stationary \\ Transformer\end{tabular}} &
			\multicolumn{2}{c}{VMD+PatchTST} \\
			\midrule
			Data & Metric & \multicolumn{1}{c}{MSE} & \multicolumn{1}{c}{sMAPE} & \multicolumn{1}{c}{MSE} & \multicolumn{1}{c}{sMAPE} & \multicolumn{1}{c}{MSE} & \multicolumn{1}{c}{sMAPE} & \multicolumn{1}{c}{MSE} & \multicolumn{1}{c}{sMAPE} \\
			\midrule
			\multicolumn{2}{c}{SP500} & 92.2142 & 0.0624 & 65.4066 & 0.028 & 11.9431 & 0.0057 & 7.6986 & 0.0042 \\
			\multicolumn{2}{c}{DJI} & 685.6056 & 0.0371 & 523.1085 & 0.0222 & 101.5294 & 0.0045 & 51.6749 & 0.0024 \\
			\multicolumn{2}{c}{SSEC} & 78.008 & 0.0234 & 140.5044 & 0.0522 & 25.476 & 0.0085 & 13.2781 & 0.0046 \\
			\multicolumn{2}{c}{FTSE} & 57.7414 & 0.0074 & 108.592 & 0.0153 & 23.038 & 0.003 & 19.913 & 0.0029 \\
			\bottomrule
		\end{tabular}
	}
	\label{table:VMD+models+ASWL}
\end{table*}

Table 5 reveals that, compared to the models in Table~\ref{table:VMD+models} , the incorporation of ASWL leads to varying degrees of improvement in the MSE and sMAPE for VMD+models. For instance, VMD+PatchTST with ASWL shows reductions in MSE of 42.28\%, 23.63\%, 25.60\%, and 13.89\% for the SP500, DJI, SSEC, and FTSE datasets, respectively. This indicates that ASWL enables the model to dynamically adjust the importance of sub-sequences during training, thereby enhancing predictive performance. Furthermore, VMD+PatchTST with ASWL achieves the lowest MSE and sMAPE across all datasets, significantly outperforming VMD+Informer with ASWL and VMD+Autoformer with ASWL. Although the VMD+Non-stationary Transformer with ASWL model also performs well, it does not match the performance of VMD+PatchTST with ASWL in terms of MSE and sMAPE. These results underscore the high effectiveness of our proposed VMD+PatchTST with ASWL in stock index forecasting, surpassing other models.

\begin{table*}[htbp]
	\centering
	\caption{Forecasting results of VMD+deep models with ASWL on IMFs of four stock index datasets}
	\resizebox{\textwidth}{!}{
		\begin{tabular}{lccccccccc}
			\toprule
			\multicolumn{2}{c}{Model with ASWL} &
			\multicolumn{2}{c}{VMD+Informer} &
			\multicolumn{2}{c}{VMD+Autoformer} &
			\multicolumn{2}{c}{\begin{tabular}[c]{@{}c@{}}VMD+Non-stationary \\ Transformer\end{tabular}} &
			\multicolumn{2}{c}{VMD+PatchTST} \\
			\midrule
			Data & Metric & \multicolumn{1}{c}{MSE} & \multicolumn{1}{c}{sMAPE} & \multicolumn{1}{c}{MSE} & \multicolumn{1}{c}{sMAPE} & \multicolumn{1}{c}{MSE} & \multicolumn{1}{c}{sMAPE} & \multicolumn{1}{c}{MSE} & \multicolumn{1}{c}{sMAPE} \\
			\midrule
			\multirow{10}{*}{SP500} & IMF0 & 88.2981 & 0.0571 & 43.7123 & 0.0185 & 4.4339 & 0.0018 & 1.9603 & 0.0012 \\
			& IMF1 & 26.657 & 0.7464 & 37.2202 & 0.7436 & 7.6252 & 0.2553 & 4.2709 & 0.1656 \\
			& IMF2 & 10.3978 & 0.689 & 15.8875 & 0.8913 & 4.5113 & 0.279 & 3.6596 & 0.2589 \\
			& IMF3 & 4.5487 & 0.8089 & 7.6768 & 0.7842 & 2.571 & 0.3245 & 2.0698 & 0.2998 \\
			& IMF4 & 3.8459 & 0.9055 & 4.1254 & 0.9149 & 1.7409 & 0.4144 & 1.4602 & 0.3754 \\
			& IMF5 & 2.7262 & 0.7941 & 2.2424 & 0.6454 & 1.5509 & 0.3978 & 1.1403 & 0.364 \\
			& IMF6 & 1.7 & 0.8096 & 2.0619 & 0.8627 & 0.9474 & 0.4374 & 0.918 & 0.451 \\
			& IMF7 & 2.0028 & 0.965 & 2.7232 & 1.1307 & 1.0061 & 0.484 & 0.9484 & 0.5092 \\
			& IMF8 & 1.2428 & 0.8415 & 1.9744 & 1.0008 & 0.7089 & 0.4682 & 0.6544 & 0.4365 \\
			& IMF9 & 1.7305 & 1.0278 & 1.7157 & 0.8351 & 0.5748 & 0.4123 & 0.5651 & 0.3952 \\
			\midrule
			\multirow{10}{*}{DJI} & IMF0 & 661.722 & 0.0355 & 363.2276 & 0.0187 & 34.1971 & 0.0021 & 13.9332 & 0.0008 \\
			& IMF1 & 62.5747 & 0.3176 & 236.523 & 0.777 & 66.641 & 0.239 & 23.875 & 0.139 \\
			& IMF2 & 44.3256 & 0.3967 & 153.7604 & 0.8636 & 33.8182 & 0.3162 & 22.4707 & 0.2256 \\
			& IMF3 & 30.7616 & 0.5078 & 64.5175 & 0.8467 & 22.2233 & 0.3575 & 13.1752 & 0.2604 \\
			& IMF4 & 31.5027 & 0.7572 & 32.8415 & 0.8476 & 14.9972 & 0.4027 & 11.6189 & 0.3439 \\
			& IMF5 & 16.2436 & 0.737 & 21.5188 & 0.7661 & 11.4901 & 0.4841 & 7.3325 & 0.3925 \\
			& IMF6 & 13.494 & 0.7005 & 30.2758 & 1.164 & 8.643 & 0.5259 & 7.356 & 0.4992 \\
			& IMF7 & 11.2981 & 0.8864 & 15.852 & 0.9395 & 6.2367 & 0.4842 & 6.4315 & 0.4909 \\
			& IMF8 & 11.1818 & 0.8183 & 14.48 & 0.9068 & 6.1468 & 0.4638 & 5.2785 & 0.4105 \\
			& IMF9 & 8.382 & 0.6602 & 29.9889 & 0.939 & 5.1909 & 0.4524 & 5.1185 & 0.4157 \\
			\midrule
			\multirow{10}{*}{SSEC} & IMF0 & 64.6236 & 0.0211 & 48.5715 & 0.0187 & 8.581 & 0.0025 & 3.06 & 0.0009 \\
			& IMF1 & 34.0917 & 0.2853 & 106.0785 & 0.8187 & 17.3159 & 0.1516 & 7.5686 & 0.1018 \\
			& IMF2 & 7.3862 & 0.3796 & 28.8456 & 0.9162 & 7.2764 & 0.3417 & 5.7119 & 0.2741 \\
			& IMF3 & 6.9483 & 0.4939 & 18.0053 & 0.9711 & 5.7538 & 0.3609 & 3.383 & 0.2644 \\
			& IMF4 & 4.5059 & 0.5522 & 6.9558 & 0.7914 & 3.7996 & 0.4489 & 2.7392 & 0.3557 \\
			& IMF5 & 3.0072 & 0.6592 & 7.6636 & 0.9386 & 2.2564 & 0.4847 & 1.9575 & 0.4227 \\
			& IMF6 & 2.9507 & 0.8171 & 4.797 & 1.0592 & 1.9129 & 0.5428 & 1.6517 & 0.4863 \\
			& IMF7 & 1.6038 & 0.6205 & 3.2885 & 0.9539 & 1.7945 & 0.5355 & 1.3355 & 0.4739 \\
			& IMF8 & 2.1158 & 0.7612 & 3.2078 & 0.9395 & 1.2329 & 0.4443 & 1.0104 & 0.4355 \\
			& IMF9 & 1.584 & 0.7436 & 1.8827 & 0.8252 & 0.973 & 0.4664 & 0.8378 & 0.3794 \\
			\midrule
			\multirow{10}{*}{FTSE} & IMF0 & 48.5952 & 0.0064 & 97.6498 & 0.0149 & 9.6133 & 0.0014 & 12.6752 & 0.0023 \\
			& IMF1 & 25.8417 & 0.5776 & 79.3896 & 0.9589 & 10.9676 & 0.2711 & 7.5401 & 0.1685 \\
			& IMF2 & 11.6413 & 0.4828 & 27.7935 & 0.7939 & 9.9045 & 0.3264 & 7.461 & 0.257 \\
			& IMF3 & 6.2662 & 0.3977 & 16.8503 & 0.856 & 6.5572 & 0.3739 & 4.8081 & 0.2995 \\
			& IMF4 & 6.6418 & 0.5669 & 11.8172 & 0.8418 & 4.6207 & 0.4339 & 3.7721 & 0.3196 \\
			& IMF5 & 6.4193 & 0.7545 & 7.8079 & 0.8954 & 3.623 & 0.5096 & 3.6036 & 0.4751 \\
			& IMF6 & 5.2092 & 0.8143 & 7.7983 & 1.0383 & 2.8666 & 0.525 & 2.7927 & 0.5295 \\
			& IMF7 & 3.788 & 0.7758 & 4.9985 & 0.9318 & 2.6176 & 0.55 & 2.2188 & 0.5161 \\
			& IMF8 & 3.1009 & 0.731 & 4.2166 & 0.9126 & 1.8087 & 0.4835 & 1.9567 & 0.4661 \\
			& IMF9 & 2.3683 & 0.6759 & 5.3718 & 1.174 & 1.3705 & 0.4377 & 1.3338 & 0.4062 \\
			\bottomrule
		\end{tabular}
	}
	\label{table:IMF10_ASWL}
\end{table*}

Table~\ref{table:IMF10_ASWL} provides a detailed analysis of the VMD+deep models with ASWL framework applied to IMFs of the same stock index datasets. The table breaks down the results by IMFs (IMF0 to IMF9), offering a nuanced perspective on the model's performance across different frequency components. The results indicate that, compared to Table~\ref{table:IMF10}, the VMD+models with ASWL framework has shown increased attention to low-frequency subsequences (IMFs 0-2), with significant reductions in both MSE and sMAPE. Specifically, VMD+PatchTST with ASWL has achieved reductions in MSE of 38.26\%, 35.55\%, and 14.20\% for IMF0-2 in the DJI dataset. This demonstrates that ASWL effectively captures the scale information of each IMF. Consequently, as the data scale increases, the improvement in prediction performance of the model with ASWL for that subsequence becomes more pronounced. Similarly, VMD+PatchTST with ASWL also excels in forecasting high-frequency IMFs, achieving the lowest MSE and sMAPE values. For instance, VMD+PatchTST recorded an MSE of 1.3338 and an sMAPE of 40.62\% for IMF9 in the FTSE dataset, showcasing its effectiveness in handling high-frequency fluctuations.

\begin{figure*}[htbp]
	\centering
	\includegraphics[width=0.6\textwidth]{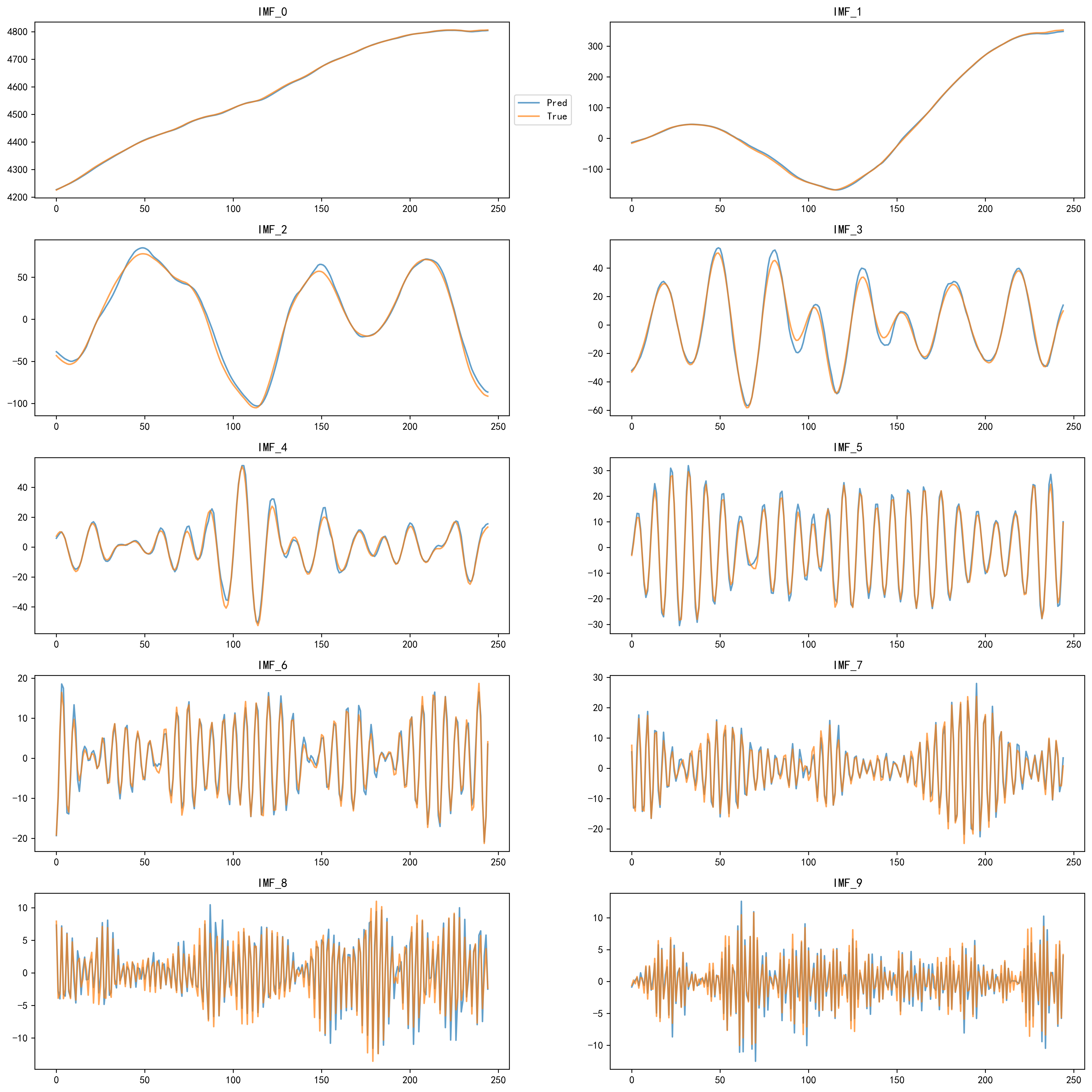}
	\caption{Forecasting results of IMFs for the fifth period SP500 using VMD+PatchTST with ASWL}
	\label{fig:imf_ASWL_pred}
\end{figure*}

Furthermore, we present the predictive plot of the VMD+PatchTST with ASWL model for the SP500 dataset at period five, as shown in Fig.~\ref{fig:imf_ASWL_pred}. This figure clearly illustrates the model's forecasting performance across different IMFs, with particularly notable accuracy in IMFs 1-3. This excellent fit is a key factor enabling the VMD+PatchTST to more accurately predict stock prices. IMFs 1-3 typically capture the main mid-frequency components of the series, and precise forecasting of these components effectively captures the primary trends in stock price movements, thereby enhancing overall prediction accuracy. For high-frequency IMFs (e.g., IMF9), the inclusion of ASWL further improves the model's performance. The results in Fig.~\ref{fig:imf_ASWL_pred} demonstrate that the prediction fluctuations for high-frequency components are better stabilized, with a significant reduction in error. This improvement indicates that ASWL not only enhances the model's ability to capture long-term trends but also optimizes the handling of short-term fluctuations and noise, leading to more stable and reliable predictions. Specifically, ASWL, through its adaptive weighting mechanism, effectively adjusts the impact of different frequency components on the final forecast, resulting in better error and fluctuation reduction when processing high-frequency data.

\subsection{Summarizations of forecasting results}

From the above analysis, we can summarize the findings as follows: (1) due to the non-stationarity and complexity of stock price series, individual models struggle to achieve satisfactory accuracy when directly learning from these sequences. (2) PatchTST model demonstrates superior performance in predicting stock index price sequences, primarily as a result of its patching mechanism. (3) combining VMD with deep models significantly reduces the prediction error for stock index prices, as measured by MSE and sMAPE, with VMD+PatchTST achieving the best results. (4) The ASWL module further improves predictive performance by incorporating scale information that is not considered by the deep model. Additionally, this module effectively reduces prediction errors in low-frequency subsequences and diminishes volatility in high-frequency subsequences, thereby optimizing the overall predictive capability of the model.

\section{Conclusion}
\label{sec:conclusion}
Forecasting stock index prices presents a complex challenge due to the inherent non-stationarity and intricate patterns of the raw price series. To address these challenges, we propose a novel forecasting framework that integrates variational mode decomposition, PatchTST, and adaptive scale-weighted layer within the established paradigm of "decomposition and integration." In our approach, the raw stock index price series is first decomposed into several IMFs using VMD, each of which exhibits more manageable characteristics compared to the original series. For each IMF, we apply PatchTST to capture and model temporal patterns effectively. The ASWL module is then employed to incorporate scale information that enhances the predictive performance. The final forecast is obtained by aggregating the results from all IMFs.

The novelty of our method lies in the integration of VMD with PatchTST and ASWL, leveraging the strengths of decomposition, temporal pattern modeling, and adaptive scale-weighted. Extensive experiments and comparative analyses validate the effectiveness and efficiency of the proposed VMD-PatchTST-ASWL framework. Future work will focus on applying this integrated approach to other multivariate time series forecasting tasks, such as energy price prediction, load forecasting, and wind speed forecasting. This will provide further insights into the model's versatility and robustness across different scenarios.

\section*{Acknowledgements}
This work is supported by the National Social Science Foundation of China [23BJY221]; the National Key Research and Development Program of China [2021QY2100].

\bibliographystyle{unsrt}
\bibliography{references}

\end{document}